# High-Precise Robot Arm Manipulation based on Online Iterative Learning and Forward Simulation with Positioning Error Below End-Effector Physical Minimum Displacement

Qu Weiming[&], Liu Tianlin[&], Luo Dingsheng[*]

* Corresponding author

[&]These authors contributed equally to this work and should be considered co-first authors.

the National Key Laboratory of General Artificial Intelligence, Key Laboratory of Machine Perception (MoE),
School of Intelligence Science and Technology, Peking University, Beijing 100871, China.
Email: {quweiming, liutl, dsluo}@pku.edu.cn

*Abstract*— **Precision is a crucial performance indicator for robot arms, as high precision manipulation allows for a wider range of applications. Traditional methods for improving robot arm precision rely on error compensation. However, these methods are often not robust and lack adaptability. Learning-based methods offer greater flexibility and adaptability, while current researches show that they often fall short in achieving high precision and struggle to handle many scenarios requiring high precision. In this paper, we propose a novel high-precision robot arm manipulation framework based on online iterative learning and forward simulation, which can achieve positioning error (precision) less than end-effector physical minimum displacement. Additionally, we parallelize multiple high-precision manipulation strategies to better combine online iterative learning and forward simulation. Furthermore, we consider the joint angular resolution of the real robot arm, which is usually neglected in related works. A series of experiments on both simulation and real UR3 robot arm platforms demonstrate that our proposed method is effective and promising. The related code will be available soon.**

## I. INTRODUCTION

Precision is a key performance indicator for robot arms and robot arm high-precision manipulation has a wide range of applications, including aerospace assembly, laser cutting, and medical puncturing. High precision enables a robot arm to be used effectively and safely, while low precision may lead to system failures and safety issues. The precision of a robot arm includes absolute positioning precision, which refers to the deviation of the actual position from the theoretical position, and repeat positioning precision, which refers to the precision of the robot arm to reach the same position. In this paper, we focus on absolute positioning precision.

Traditional methods used for improving the precision of robot arms is the error compensation method, which involves offsetting an error by applying a value that is the opposite of the error [1-3]. However, the error compensation method has low robustness and poor adaptability. Learning-based methods have been widely used in recent years due to their advantages in flexibility and adaptability, but current researches show that they are not capable of handling many scenarios with high-precision requirements [4][5]. The FetchReachEnv in OpenAI Gym [6] [7], which is widely used for robot arm reaching, only sets the default precision threshold to 5 cm. Besides, in a recent study by Aumjaud et al. [8], multiple precision thresholds were used to compare the performance of different reinforcement learning methods, but the lowest precision threshold was only set to 5 mm. One possible reason for the low precision of learning-based methods is the neglect of the sparsity problem of training data for inverse models. In addition, most learning-based methods ignore the joint angle resolution and assume that all joint angles can be executed. However, for a real robot arm, the joints cannot perform arbitrary angles due to physical limitations.

In this paper, we propose a high-precision manipulation framework for robot arms based on online iterative learning and forward simulation. To address the sparse problem of inverse model training data, we pre-train a global "average" inverse model and then perform online learning of the inverse model for desired target positions to achieve high precision. Drawing on the idea of mental simulation [9] and combining it with our previous work, embodied self-supervised learning (EMSSL) [10], we develop multiple strategies to better combine online iterative learning and forward simulation, and improve system performance by parallelizing them. At the same time, we use relative position instead of absolute position, which greatly reduces the problems in terms of visual perception errors that existed in our previous work [10]. Furthermore, due to the property of the proposed framework that the forward model does not need to be derivable, joint angular resolution is considered in this paper. A series of experiments on both simulation and real UR3 robot arm platforms demonstrate that our proposed method is effective and promising, where the positioning error (precision) is less than the end-effector physical minimum displacement.

In summary, the main contributions of this paper are summarized as follows:

- A novel high-precision manipulation framework for robot arm based on online iterative learning and forward simulation is proposed, which enables the positioning error below the end-effector physical minimum displacement.

- Multiple strategies are parallelized to better combine online iterative learning and forward simulation, which further improves the system performance.

- Relative position is introduced instead of absolute position to effectively reduce visual perception errors in the real-world environment.

- Joint angular resolution of the real robot arm is considered, which is usually neglected in related works.

## II. RELATE WORKS

### A. Traditional Methods

Traditional methods used to improve the precision of the robot arm is the error compensation methods, which use measurement tools and numerical algorithms to eliminate error after the assembly is completed and adjust preset parameters until the robot meets precision requirements [1-3]. Error compensation methods can be classified into two categories based on differences in principle and ideas: offline calibration methods and online compensation methods.

(1) Offline calibration methods

The main idea behind offline calibration methods is to measure the positioning error of several joint configurations in the robot workspace and establish a robot kinematic error model to identify the robot kinematic parameter error or establish an error mapping in the robot's Cartesian space or joint space. The obtained error model or error mapping is then pre-programmed into the compensation algorithm to compensate for the positioning error.

A commonly used approach is kinematic calibration [11-13], which involves establishing a mathematical model that describes the geometric characteristics and kinematic performance of the robot arm. Multi-point positions of the robot end-effectors are then measured, and the joint kinematic parameter errors are identified and brought into the robot arm kinematic model to minimize the residuals between the estimated and actual values of the robot positioning errors. One of the main problems with kinematic calibration is that many factors affecting robot positioning errors are coupled with each other, making it difficult to construct an accurate model that includes all error sources. To solve this problem, researchers focused on constructing a general model to predict errors by mapping positioning errors to joint angles and constructing the corresponding error models [14-17]. However, these approaches require high-precision external detection equipment to measure end positioning errors, limiting their application in some cases. To address this shortcoming, researchers developed physically constrained calibration approaches [18-20], which involve constructing constraint equations by contacting the end of the robot arm with physical constraints, such as balls, planes, and single points, based on which the error model is built.

However, offline calibration methods cannot compensate for dynamic errors in the actual manipulation of the robot arm. Moreover, offline calibration methods depend highly on the repetitive positioning precision of the robot arm, and unfortunately, the multi-directional repetitive positioning precision of the robot arm is usually poor.

(2) Online compensation methods

Online compensation approaches can be a good solution to these limitations. They use external high-precision measurement devices to provide real-time feedback to the motion of the robot arm, allowing the robot to continuously adjust its pose until the target state is achieved during the working process.

A classical online compensation method is visual servoing, which uses feedback information from vision sensors to guide the robot [21-24]. One approach is image-based visual servoing (IBVS) [25-27], first proposed by Sanderson et al. in 1983 [28]. The robot compares the currently seen image with the target image and calculates the corresponding amount of manipulation based on the difference between them. However, IBVS usually lacks depth information, leading to a large position error. Another approach is position-based visual servoing (PBVS) [29-32], which first calculates the position of the target object and then calculates the corresponding manipulation amount based on the difference between the current position of the robot and the target position. Although PBVS can effectively reduce the position error, the pose error is usually large. Researchers have proposed combining IBVS and PBVS to reduce pose error by using image information and reduce position error by using coordinate information. However, it is difficult to be used for high-frequency real-time manipulation due to the homography matrix decomposition in the servoing process.

Although error compensation methods can achieve high-precision manipulation for robot arm, existing methods often lack robustness to uncertainty perturbations, leading to degradation of the robot arm's precision. Error compensation methods are also more costly and less efficient.

### B. Learning-based methods

To address the shortcomings of traditional methods, learning-based methods have received extensive attention in recent years. Jiang et al. [33] realized the manipulation of a piercing robot based on particle swarm optimization and a BP neural network, achieving a final precision of 1 mm, which meets the requirements of medical piercing robots. Fan et al. [34] combined supervised learning and reinforcement learning to propose a high-precision industrial assembly learning framework with a final precision of 0.2 mm. Luo et al [35] proposed precision-based continuous curriculum learning (PCCL) for robot arm reaching tasks. Unlike traditional methods that use a fixed precision threshold, the precision threshold of PCCL is gradually adjusted during the training process. In 2021, Aumjaud et al [8] designed a benchmark experimental environment, rl_reach [36], to compare the performance of various model-free reinforcement learning methods on the task of robot arm reaching. This environment contains various reinforcement learning methods, such as proximal policy optimization (PPO) [37], soft actor-critic (SAC) [38], and twin delayed deep deterministic policy gradient (TD3) [39]. However, in their baseline experiments, the precision threshold is set to a minimum of only 5 mm.

In general, learning-based approaches have significant advantages in terms of flexibility and adaptability, and thus have become a hot topic in robot arm high-precision manipulation. However, supervised learning-based approaches suffer from the problem of sparse inverse model training data, which makes it difficult to train an inverse model with high precision for all positions in the workspace. Reinforcement learning-based approaches often result in poor precision and require multiple interactions with the environment, making them less efficient. Additionally, the robot arm is often used in conjunction with vision sensors, and the use of absolute position of the target can result in large visual perception errors, leading to poor final precision.

## III. METHODOLOGY

In this section, we will first highlight the benefits of using relative position and propose an algorithm for learning the inverse model of a robot arm based on relative position, building upon our previous work [10]. Subsequently, we will address the challenge of sparse inverse model training data and propose a high-precision manipulation framework for robot arm based on online iterative learning and forward simulation. Finally, we will develop various strategies to effectively combine online iterative learning and forward simulation and parallelize them for further improving the system performance.

### A. Relative Position Based Inverse Model EMSSL

Our previous work [10] takes the absolute position of the target object as the input of the inverse model. Although the error of the inverse model itself is small, the final error is still large due to the presence of visual perception error. Using relative position can reduce the error of visual perception, which can also avoid the cumulative error caused by the absolute position computing which requires multiple coordinate transformations. The relative position can be obtained from a binocular camera at the end of the robot arm. Although the binocular camera can also be installed in other places, such as some humanoid robots have a binocular camera on the head, installing the binocular camera at the end of the robot arm can make the visual perception error smaller, because the closer to the target object, the higher the pixel resolution, and the smaller the visual perception error, as is shown in Fig. 1. In this paper, we obtain the relative position of the target object from the end of the robot arm, i.e., the position of the target object in the end camera coordinate system, based on the pinhole camera model and binocular parallax principle [40]. Fig. 2 shows the diagram of binocular parallax principle. According to the similar triangle principle:

$$\frac{b-(u_l-u_r)}{b} = \frac{p_z-f}{p_z} \quad (1)$$

that is

$$p_z = \frac{fb}{u_l-u_r} \quad (2)$$

where $u_l - u_r$ is the parallax. $f$ is the focal distance, $b$ is the baseline, and $p_z$ is the depth.

From equation (2), we can see that the depth is inversely proportional to the parallax, which also shows that the measurement of a near object is more accurate than that of a distant object. According to the pinhole camera model, $p_x$ and $p_y$ can be obtained as follows:

$$\begin{cases} p_x = \frac{(u_l-c_x)p_z}{f_x} \\ p_y = \frac{(v_l-c_y)p_z}{f_y} \end{cases} \quad (3)$$

where $c_x, c_y, f_x, f_y$ are camera intrinsic parameters.

Unlike [10], the input of the inverse model in this paper is the robot arm state and the relative position, and the output is the joint angle variation. The input of the forward model is the robot arm state and the joint angle variation, and the output is the predicted relative position. The relative position based inverse model embodied self-supervised learning algorithm is shown in Algorithm 1.

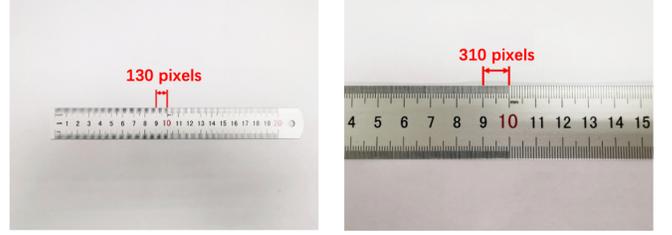

(a) far from the target object    (b) close to the target object

Fig.1. The pixel resolution of the same object (ruler) at different distances. The image sizes of (a) and (b) are the same. As can be seen, the number of pixels in 10 mm is 130, i.e., a resolution of 13 pixels/mm, when the object is far away from the camera; while the number of pixels in 10 mm is 310, i.e., a resolution of 31 pixels/mm, when the object is close to the camera.

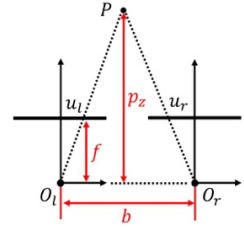

Fig.2. The diagram of binocular parallax principle, where $P$ is the target point, $O_l$ is the aperture center of the left camera, $O_r$ is the aperture center of the right camera, $(u_l, v_l)$ are the pixel coordinates of $P$ in the left camera, and $(u_r, v_r)$ are the pixel coordinates of P in the right camera (only one direction is indicated in this figure, so $v_l$ and $v_r$ are not labeled here).

### B. A Robot Arm High-precision Manipulation Framework

A problem that is often overlooked is the sparsity of the training data for the inverse model, which makes it difficult to train an inverse model with high precision for all target positions in the robot arm workspace. Assuming that the robot arm has $n$ degrees of freedom, the joint angle resolution is $\alpha_i$, and the joint angle range is $r_i$, $i = 1, 2, \ldots, n$, then the data space size of the inverse model is $\prod_{i=0}^{n} \frac{r_i}{\alpha_i}$. Taking the UR3 robot arm used in this paper as an example, it is a 6 degree-of-freedom robot arm with a joint angle resolution of 0.01°. If the joint angle range settings in Table I, the data space size of the inverse model is $5.3 \times 10^{23}$, which is a large amount of data. The amount of data used in this paper to train the inverse model is only $10^5$, which is a small amount of data relative to the entire data space size. The sparse data volume also makes it difficult to fit a neural network to an inverse model with high global precision, especially for "unseen" target positions.

TABLE I. ANGLE RANGE OF EACH JOINT

| Joint | Angle Range (°) |
|---|---|
| Base | [-90, 0] |
| Shoulder | [-180, -90] |
| Elbow | [-90, 0] |
| Wrist 1 | [-135, -45] |
| Wrist 2 | [45, 135] |
| Wrist 3 | [90, 180] |

**Algorithm 1** Relative Position Based Inverse Model EMSSL

**Input:** Forward Model FM, Inverse Model IM, Unlabeled dataset of states and relative positions of the robot arm $\mathcal{U} = \{(s^{(i)}, \Delta p^{(i)})\}_{i=1}^{N}$, Maximum number of iterations $T$, Epochs $E$, Number of batches in a round of epoch for dataset (for example $\mathcal{X}$) $N_{B\mathcal{X}}$, Number of small batch samples for inference $M_R$, Number of small batch samples for training $M_T$, Number of parallel computation threads $K$, Learning rate $\eta$

1:    Initialize the parameters $\boldsymbol{\theta}$ of the inverse model IM randomly
      Initialize the sample dataset: $\mathcal{D} \leftarrow \emptyset$
      Initialize the joint angle variation dataset of IM inference: $\mathcal{Q} \leftarrow \emptyset$
      Initialize the relative position dataset of FM computation: $\mathcal{P} \leftarrow \emptyset$
      Initialize the robot arm state dataset: $\mathcal{S} \leftarrow \emptyset$
2:    **for** $t = 1 \ldots T$ **do**
3:       \*\* ***Data Sampling*** \*\*
4:       Empty the dataset $\mathcal{D}, \mathcal{Q}, \mathcal{P}, \mathcal{S}$
5:       **for** $n = 1,2 \ldots N_{B\mathcal{U}}$ **do**
6:          Sample small batches from $\mathcal{U}$: $\mathcal{B} \leftarrow \{(s^{(m)}, \Delta p^{(m)})\}_{m=1}^{M_R}$
          Separate $\mathcal{B}$ to get $\mathcal{S}_n$ and $\mathcal{P}_n$
7:          IM batch inference: $\mathcal{Q}_n \leftarrow IM(\mathcal{S}_n, \mathcal{P}_n)$
8:          Update the dataset: $\mathcal{Q} \leftarrow \mathcal{Q} \cup \mathcal{Q}_n$, $\mathcal{S} \leftarrow \mathcal{S} \cup \mathcal{S}_n$
9:       **end for**
10:      $j \leftarrow 0$
11:      **repeat**
12:         According to the number of threads $K$, pop the state and the joint angle variation in order:
           $\mathcal{S}_j, \mathcal{Q}_j = \{(s_{1+j}, \Delta q_{1+j}), (s_{2+j}, \Delta q_{2+j}), \ldots, (s_{k+j}, \Delta q_{k+j})\}$
13:        FM parallel computation: $\mathcal{P}_j = FM(\mathcal{S}_j, \mathcal{Q}_j)$
14:        Update the dataset: $\mathcal{P} \leftarrow \mathcal{P} \cup \mathcal{P}_j$
15:        $j = j + 1$
16:      **until** the data in $\{\mathcal{S}, \mathcal{Q}\}$ have been computed
17:      Update the sample dataset: $\mathcal{D} \leftarrow \{\mathcal{S}, \mathcal{Q}, \mathcal{P}\}$
18:      \*\* ***Model Training*** \*\*
19:      **for** $e = 1 \ldots E$ **do**
20:         **for** $n = 1,2 \ldots N_{B\mathcal{D}}$ **do**
21:            Sample small batches from $\mathcal{D}$:
              $\mathcal{B} \leftarrow \{(s^{(m)}, \Delta q^{(m)}, \Delta p^{(m)})\}_{m=1}^{M_T}$
22:            $L(\boldsymbol{\theta}) = \frac{1}{M_T}\sum_{i=1}^{M_T}\left(\Delta q^{(m)} - IM((s^{(m)}, \Delta p^{(m)}))\right)^2$
23:            Update $\boldsymbol{\theta}$ with GD: $\boldsymbol{\theta} \leftarrow \boldsymbol{\theta} - \eta\nabla_{\boldsymbol{\theta}}L(\boldsymbol{\theta})$
24:         **end for**
25:      **end for**
26:    **end for**

Therefore, in this paper, we propose pre-training a global "average" inverse model first, instead of directly training an inverse model with high precision for all target positions in the workspace. Subsequently, online learning of the inverse model is performed based on the required target positions to achieve high precision. The high-precision robot arm manipulation framework proposed in this paper is depicted in Fig. 3. Firstly, Algorithm 1 is used to obtain a pre-trained inverse model. However, due to the sparsity of the training data of the inverse model, the precision of the pre-trained inverse model is usually unsatisfactory. Therefore, for a new target, the inverse model continues to learn online iteratively based on the previously pre-trained inverse model until the precision reaches the threshold. In this process, the forward simulation module is used to solve the problem of falling into local minima during online iterative learning and accelerate learning.

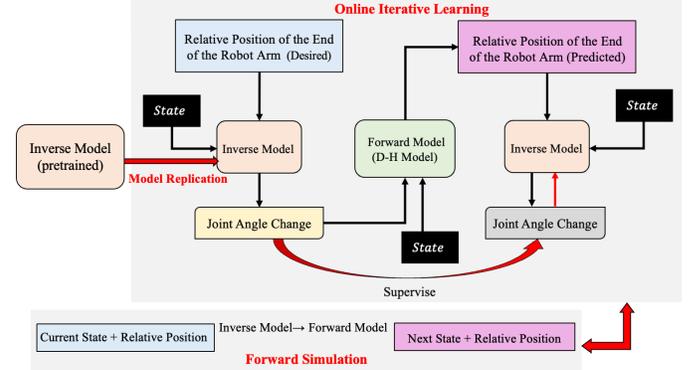

Fig.3. The framework of robot arm high-precision manipulation based on online iterative learning and forward simulation, which consists of three modules: model replication, online iterative learning and forward simulation.

(1) Model Replication

The idea of model replication is similar to that of Model-Agnostic Meta-Learning (MAML) [41]. First, an inverse model is pre-trained using Algorithm 1, after which the model parameters of this inverse model remain unchanged and it can be considered as an "average" inverse model. For a new target, the pre-trained inverse model is first replicated, and then the replicated inverse model is used for online iterative learning. Model replication avoids overfitting of the pre-trained inverse model to a single target, resulting in a lower average cost.

(2) Online Iterative Learning

Unlike inverse model learning during pretraining, online iterative learning is more purposeful. It targets the currently reaching target positions, while inverse model learning during pretraining iterates over the entire dataset. Since the error of the pre-trained inverse model is relatively small, a smaller initial learning rate should be chosen for online iterative learning. The algorithm for online iterative learning is shown in Algorithm 2.

(3) Forward Simulation

Forward simulation is based on the mental simulation mechanisms of human cognition [9]. The process of forward simulation is as follows: First, the relative position of the end of the robot arm can be obtained for a given target. Then, the relative position and the current state of the robot arm are input into the inverse model to deduce the corresponding variation in joint angles. Finally, the forward model is used to predict the new state and relative position of the robot arm after execution. In this process, the robot arm does not actually perform the action. Forward simulation allows the inverse model to learn iteratively in a smaller area, thus speeding up the convergence of the inverse model. It can also help the inverse model avoid falling into local minima during online iterative learning. Furthermore, since the robot can be simulated in the "virtual brain" without actual execution, the sampling efficiency is greatly improved.

---

**Algorithm 2**    Online Iterative Learning

---

**Input:** Forward Model FM, Inverse Model IM, Current State $s_{cur}$, Relative position of the robot arm $\Delta p_{cur}$, Maximum number of iterations $T$, Learning rate $\eta$, Epochs $E$, Threshold $THRESHOLD$
**Output:** The best inverse model parameters $\theta_{best}$, the precision $pr_{best}$, and the joint angle variation $\Delta q_{best}$

1:   Initialize $pr_{best} \leftarrow \inf$, $q_{best} \leftarrow 0$, $\theta_{best}$
2:   **for** $t = 1 \ldots T$ **do**
3:      IM inference: $\Delta q \leftarrow IM(s_{cur}, \Delta p_{cur})$
4:      FM computation: $\Delta P = FM(s_{cur}, \Delta q)$
5:      **for** e= 1,2 ... $E$ **do**
6:          Compute loss: $L(\theta) = (\Delta q - IM(s_{cur}, \Delta P))^2$
7:          Update $\theta$ with GD: $\theta \leftarrow \theta - \eta \nabla_\theta L(\theta)$
8:      **end for**
9:      Compute the joint angle variation: $\Delta q_{new} \leftarrow IM(s_{cur}, \Delta p_{cur})$
10:     Compute the prediction position: $\Delta p_{new} \leftarrow FM(s_{cur}, \Delta q_{new})$
11:     Compute the precision: $pr \leftarrow \|\Delta p_{new}\|_2$
12:     **if** $pr < pr_{best}$ **then**
13:         $pr_{best} \leftarrow pr$
14:         $\Delta q_{best} \leftarrow \Delta q_{new}$
15:         $\theta_{best} \leftarrow \theta$
16:     **end if**
17:     **if** $pr_{best} < THRESHOLD$ **then**
18:         **return** $\theta_{best}, pr_{best}, \Delta q_{best}$
19:     **end if**
20:   **end for**
21:   **return** $\theta_{best}, pr_{best}, \Delta q_{best}$

---

### C. Different High-precision Manipulation Strategies

In order to better combine online iterative learning and forward simulation for high-precision manipulation, different strategies are illustrated in this section. The basic strategy is to use only online iterative learning. After combining with forward simulation, two different high-precision manipulation strategies are available, as is shown in Fig.4. For strategy 1, if the precision after online iterative learning cannot reach the threshold, forward simulation will be performed. Strategy 2 performs forward simulation before online iterative learning. In practice, Strategy 1 and Strategy 2 are used in parallel. For a target location, both strategies compute the solution for a target location. If the earliest solution returned meets the requirement, it is chosen. If not, the solution with the best precision among the two returned is chosen.

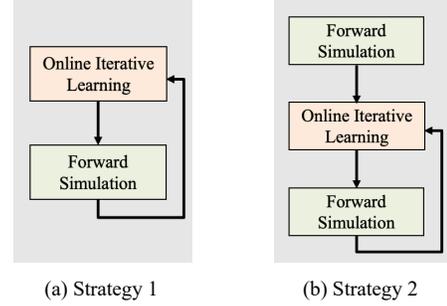

(a) Strategy 1        (b) Strategy 2

Fig.4. Two different high-precision manipulation strategies for combining online iterative learning and forward simulation, including strategy 1 "Loop (online iterative learning → forward simulation)" and strategy 2 "Forward simulation → Loop (online iterative learning → forward simulation)".

## IV. EXPERIMENTS

### A. Environmental Setup

In this paper, we developed a high-precision robot arm manipulation experiment platform, as shown in Fig. 5. The platform consists of a UR3 robot arm with a thin needle at the end, a ZED Mini binocular camera, and an Azure Kinect DK RGB-D camera. The robot arm is fixed on a table with dimensions of 120 cm × 80 cm × 90 cm. The binocular camera is mounted at the end of the robot arm, diagonally above the needle, to obtain the relative position of the needle. The RGB-D camera is mounted above the base of the robot arm and is primarily used to identify and locate the target object. Additionally, we built a simulation platform that corresponds to the real platform.

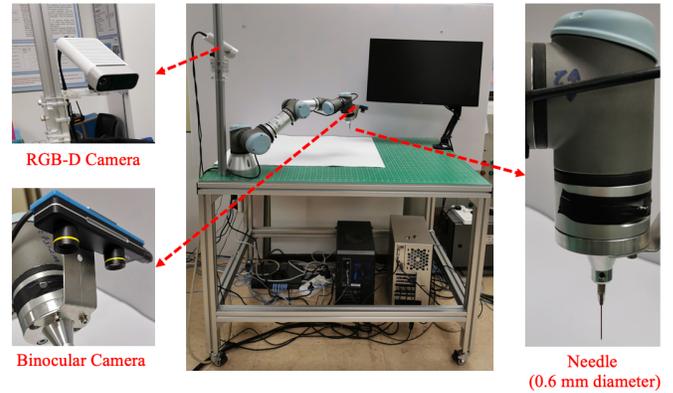

Fig.5. A high-precision robot arm manipulation experiment platform based on a UR3 robot arm, a ZED Mini binocular camera, and an Azure Kinect DK RGB-D camera.

We restrict the angle range of each joint of the robot arm according to its working space size, as shown in Table I. To train the inverse model, we first sampled a joint angle $q$ in the joint space, and then sampled the joint angle variation $\Delta q$. We calculated the variation of end position $\Delta p$ based on these samples, and used $(q, \Delta p)$ as the input of the inverse model. 140,000 data were collected in the simulation environment, of which 100,000 were used for training and the remaining 40,000 for testing. Since the maximum relative distance, i.e., the distance of the end of the robot arm from the target, varies at different ranges, we also explored the range of the joint angle variation $\Delta q$. Table II shows that the maximum relative distance is 22.26 cm when the range of joint angle is [-10°, 10°], which satisfies the requirements of this paper.

TABLE II. MAXIMUM RELATIVE DISTANCE UNDER DIFFERENT JOINT ANGLE RANGE

| Joint Angle Variation Range (°) | Maximum Relative Distance (cm) |
|---|---|
| [-1, 1] | 2.09 |
| [-5, 5] | 10.45 |
| [-10, 10] | 22.26 |

*B. Evaluation on Relative Position Based Inverse Model*

The hyperparameters for learning the relative position-based inverse model are presented in Table III. However, this model may fail to learn if its output remains the same, regardless of the input. Although the loss in Algorithm 1 is zero, the output is incorrect. To address this issue, we added random noise to the joint angle variation in the inverse model output. The random noise follows a Gaussian distribution with a diagonal covariance matrix and a main diagonal element of 0.07. After each iteration, the Gaussian noise decays with a factor of 0.001.

TABLE III. HYPERPARAMETERS SETTINGS

| Hyperparameters | Value |
|---|---|
| Activation | ReLU (Hidden Layers) |
| | Sigmoid (Output Layers) |
| Optimizer | Adam |
| Learning rate | 0.0015 |
| Batch size (Inference) | 512 |
| Batch size (Training) | 128 |
| Number of parallel computation threads | 6 |
| Maximum number of iterations | 200 |
| Epoch | 10 |
| Number of network layers | 6 |
| Gaussian Noise Parameters | 0.07 |
| Noise Decay factor | 0.001 |
| Network Type | Fully connected neural network (FCNN) |
| Network Structure | 9 → 1024 → 512 → 256 → 128 → 6 |

We chose the Direct Regression Learning (DRL) based on relative position and EMSSL based on absolute position as baseline methods. The precision, defined as the distance between the end of the robot arm and the target, was used as the evaluation indicator. The experimental results are presented in Table IV. Despite having a small range of joint angle variation ([-10°, 10°]), there is still a non-convexity problem, which leads to poor precision. While using relative position improves the precision compared to our previous work on absolute position-based inverse model learning [10], it is still not satisfactory. In the next section, we will present the experimental results obtained using the high-precision manipulation framework proposed in this paper.

TABLE IV. PERFORMANCE COMPARISON OF DIFFERENT METHODS

| Method | Precision (mm) |
|---|---|
| DRL (Relative position) | 5.55 |
| EMSSL (Relative position) | 2.80 |
| **EMSSL (Relative position)** | **1.97** |

*C. Evaluation on High-precision Manipulation Framework*

The initial learning rate of the inverse model is 0.0001 for online iterative learning. The precision threshold for high-precision reaching is set to the end-effector physical minimum displacement for different joint angle resolutions. A common high-precision manipulation method is chosen as the baseline method: iteratively reaching the target based on relative position without updating the inverse model [31][42]. The joint angle resolution is 0.01°, which is the default joint angle resolution of the UR3 robot arm. The experimental results are shown in Table V. As can be seen from the table, the result of the baseline method is similar to the result of the inverse model after pre-training (1.97 mm), while the precision of our proposed method is relatively high, which demonstrates the importance of online iterative learning for improving precision in robot arm manipulation.

TABLE V. PERFORMANCE COMPARISON OF DIFFERENT METHODS

| Method | Precision (mm) |
|---|---|
| without Inverse Model Updating (Baseline) | 1.84 |
| **Online Iterative Learning (We Proposed)** | **0.02** |

However, if the end of the robot arm has already reached the target with high precision, there may be a situation where the end joint moves forward or backward by a minimum displacement, the precision does not improve further. Based on this consideration, in order to investigate the best precision (average) for our proposed high-precision manipulation method, we set the precision threshold to be half of the minimum displacement of the end joint's movement. As is shown in Table VI, three different joint angle resolutions are investigated. Experimental results demonstrate that the average precision is less than the end-effector physical minimum displacement (close to half of it) for different joint angle resolutions.

For the experiments conducted on the real UR3 robot arm platform, the objective was to manipulate the robot arm such that the needle mounted at its end could reach the intersection of two straight lines located at the center of a rectangular block. To ensure accurate measurement, the joint angle resolution of the UR3 robot arm was set to 1°, which corresponded to a minimum displacement of 2.91 mm. The actual precision was measured using a vernier caliper with a precision of 0.01 mm. A total of 200 sets of data were tested in this paper, and the precision (average) achieved was 1.88 mm. Among these, 196 sets had a precision that was less than the minimum end joint displacement of 2.91 mm. Fig. 6 shows a sample of the experimental results.

TABLE VI. PERFORMANCE AND MINIMUM DISPLACEMENT OF THE END JOINT WITH DIFFERENT JOINT ANGLE RESOLUTION

| Joint Angle Resolution (°) | Minimum Displacement (mm) | Half of the Minimum Displacement (mm) | Precision (mm) |
|---|---|---|---|
| 0.01 | 0.03 | 0.015 | 0.016 |
| 0.1 | 0.29 | 0.145 | 0.151 |
| 1 | 2.91 | 1.455 | 1.533 |

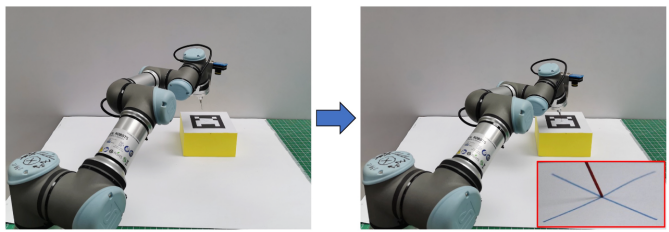

(a) Before High-precision Manipulation  (b) After High-precision Manipulation

Fig.6. A sample of the experimental results for robot arm high-precision manipulation. In this figure, before high-precision manipulation, the relative distance is 22.91mm, while the relative distance is 1.49mm after high-precision manipulation, which is less than the minimum end joint displacement of 2.91mm.

## D. Evaluation on High-precision Manipulation Strategies

In this section, the performance of different high-precision manipulation strategies, as illustrated in section III-C, is compared for three different joint angle resolution rates (0.01°, 0.1°, and 1°). In addition to precision, several other evaluation indicators are used, including Success Rate, which measures the percentage of positioning error that is less than the end-effector physical minimum displacement (precision threshold); Worst Precision, which measures the worst precision in a single reaching process; and Computation Time, which measures the computation time when different high-precision manipulation strategies are implemented.

The experimental results, as shown in Table VII, indicate that forward simulation can further improve the performance and speed of online iterative learning. Furthermore, the two strategies developed in parallel can effectively improve the success rate and speed up the computation time of the high-precision manipulation process, potentially leading to an improvement in precision as well. At the same time, the worst precision of the parallel strategy does not exceed twice the end-effector physical minimum displacement.

## V. CONCLUSION

In this paper, we propose a novel high-precision manipulation framework for the robot arm based on online iterative learning and forward simulation. Online iterative learning addresses the sparsity problem of inverse model training data, while forward simulation overcomes local minima and improves the speed of online iterative learning. We parallelize two high-precision manipulation strategies to better combine online iterative learning and forward simulation, which greatly improve the success rate (from 71.04% to 96.50%) and significantly reduced the single worst precision (from 8.17 mm to 0.04 mm). Additionally, we consider the joint angle resolution of the real robot arm, which is usually be ignored. Experimental results show that our proposed method achieves a final positioning error (precision) less than the end-effector physical minimum displacement (on the simulation platform, the positioning error is 0.02 mm for a joint angular resolution of 0.01°, and on the real robot arm, with a joint angular resolution limited to 1°, the positioning error is 1.88 mm, which is less than the end-effector physical minimum displacement of 2.91 mm). Our results demonstrate that our framework is effective and promising in achieving robot arm high-precision manipulation.

TABLE VII. PERFORMANCE COMPARISON WITH DIFFERENT STATRGIES AND DIFFERENT JOINT ANGLE RESOLUTION

| Method | Precision (mm) | Success Rate (%) | Worst Precision (mm) | Computation Time (s) | Joint Angle Resolution (°) |
|---|---|---|---|---|---|
| Basic Strategy | 0.04 | 71.04 | 8.17 | 1.18 | |
| Strategy 1 | **0.02** | 85.50 | 0.14 | 0.91 | 0.01 |
| Strategy 2 | **0.02** | 80.30 | 0.70 | 1.13 | |
| **Parallel Strategy** | **0.02** | **96.50** | **0.04** | **0.76** | |
| Basic Strategy | 0.24 | 83.30 | 7.44 | 0.81 | |
| Strategy 1 | **0.22** | 93.20 | 0.47 | 0.59 | 0.1 |
| Strategy 2 | 0.23 | 88.00 | 0.48 | 0.84 | |
| **Parallel Strategy** | **0.22** | **98.20** | **0.45** | **0.47** | |
| Basic Strategy | 2.05 | 76.95 | 6.21 | 0.65 | |
| Strategy 1 | 1.80 | 90.40 | 5.38 | 0.54 | 1 |
| Strategy 2 | 1.86 | 84.70 | 5.10 | 0.80 | |
| **Parallel Strategy** | **1.73** | **98.70** | **4.02** | **0.39** | |

ACKNOWLEDGMENT

The work is supported in part by the National Natural Science Foundation of China (No. 62176004, No. U1713217), Intelligent Robotics and Autonomous Vehicle Lab (RAV), and the Fundamental Research Funds for the Central Universities.